\documentclass{article}

\usepackage[preprint,nonatbib]{neurips_2024}

\PassOptionsToPackage{numbers, compress}{natbib}

\usepackage{natbib}
\usepackage{tabularx}
\usepackage[utf8]{inputenc} 
\usepackage[T1]{fontenc}    
\usepackage{hyperref}       
\usepackage{url}            
\usepackage{booktabs}       
\usepackage{amsfonts}       
\usepackage{nicefrac}       
\usepackage{microtype}      
\usepackage{xcolor}         
\usepackage{times}
\usepackage{latexsym}
\usepackage{graphicx}
\usepackage{subfig}
\usepackage{multirow}
\usepackage{makecell}
\usepackage{comment}
\usepackage{url}
\usepackage{booktabs}
\usepackage{amsmath}

\DeclareMathOperator*{\argmin}{arg\,min}

\title{\Large Improving General Text Embedding Model: Tackling Task Conflict and Data Imbalance through Model Merging}

\author{
Mingxin Li\textsuperscript{\rm 1},
Zhijie Nie\textsuperscript{\rm 1},
Yanzhao Zhang,
Dingkun Long,
\\ \textbf{
Richong Zhang\textsuperscript{\rm 1},
Pengjun Xie
}\\
\textsuperscript{\rm 1}CCSE, School of Computer Science and Engineering, Beihang University
\\
\texttt{\{limx,niezj,zhangrc\}@act.buaa.edu.cn}
}

\begin{document}

\maketitle

\begin{abstract}
Text embeddings are vital for tasks such as text retrieval and semantic textual similarity (STS). Recently, the advent of pretrained language models, along with unified benchmarks like the Massive Text Embedding Benchmark (MTEB), has facilitated the development of versatile general-purpose text embedding models. Advanced embedding models are typically developed using large-scale multi-task data and joint training across multiple tasks. However, our experimental analysis reveals two significant drawbacks of joint training: 1) \textbf{Task Conflict}: Gradients from different tasks interfere with each other, leading to negative transfer. 2) \textbf{Data Imbalance}: Disproportionate data distribution introduces biases that negatively impact performance across tasks. To overcome these challenges, we explore model merging—a technique that combines independently trained models to mitigate gradient conflicts and balance data distribution. We introduce a novel method, \textbf{Self Positioning}, which efficiently searches for optimal model combinations within the interpolation space of task vectors using stochastic gradient descent. Our experiments demonstrate that Self Positioning significantly enhances multi-task performance on the MTEB dataset, achieving an absolute improvement of 0.7 points. It outperforms traditional resampling methods while reducing computational costs. This work offers a robust approach to building generalized text embedding models with superior performance across diverse embedding-related tasks.
\end{abstract}

\section{Introduction}
\label{sec:introduction}
Text embedding is a crucial aspect in the fields of Information Retrieval (IR) and Natural Language Processing (NLP). Embedding models represent text in high-dimensional spaces to capture semantic information, making them applicable to various downstream tasks, including text retrieval~\cite{dpr-paper}, semantic textual similarity (STS)~\cite{gao_simcse_2021}, and text classification~\cite{tc-using-embedding-survey}, among others. Initially, text embedding models for different tasks were mostly trained independently. Recently, with the emergence of pretrained language models, such as BERT~\cite{devlin_bert_2019} and GPT3~\cite{gpt3}, and unified multi-task evaluation benchmarks like the Massive Text Embedding Benchmark (MTEB)~\cite{muennighoff_mteb_2023}, there has been significant progress in training general text embedding models~\cite{openai-embedding,wang2022text,bge-m3-paper,mgte-zhang}. These general text embeddings offer greater usability and broader applicability compared to models previously trained for specific tasks.

\begin{figure}[t]
    \centering
    \includegraphics[width=.5\columnwidth]{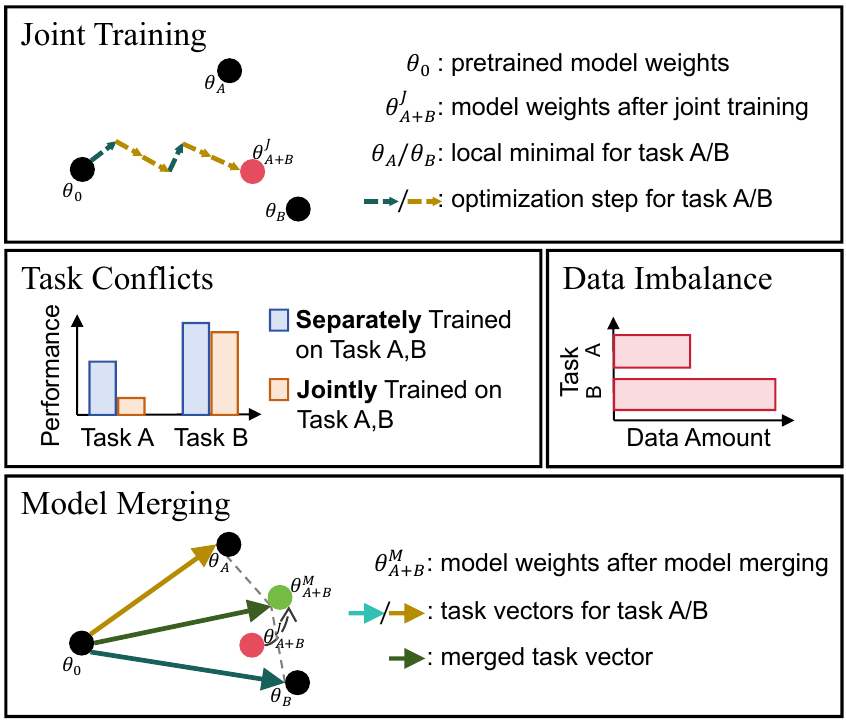}
    \caption{During joint training across multiple tasks, model weights fluctuate towards local minima for each task, which leads to two primary issues: 1) Gradient conflicts reduce performance; 2) Data imbalances result in unequal optimization frequencies, biasing the final weights towards more prevalent tasks. Model merging mitigates these challenges by isolating task training and achieving a more balanced weight position.}
    \label{fig:teaser}
\end{figure}

Recently, State-of-the-Art general text embedding models predominately adopt a dual-tower training architecture~\cite{dpr-paper}. This architecture encodes the query and document of a text pair into continuous vectors using pretrained language models, while the training objective employs contrastive learning. To ensure that the resulting embedding models possess strong domain generalization and multi-task adaptability, training typically involves collecting large-scale, multi-task relevant text pairs (often exceeding millions in quantity). Additionally, some approaches gather billions of weakly supervised text pairs from public web resources (e.g., query-answer pairs from Q\&A communities) to enhance domain generalization and robustness. Despite the effectiveness of this mixed multi-task contrastive learning approach, we have identified two significant issues that can impact model performance. The first is \textbf{Task Conflict}. Previous study~\cite{yu_gradient_2020} on multi-task learning suggests that when concurrently training on multiple tasks, gradients from different tasks may conflict, leading to the Negative Transfer Phenomenon, where a jointly trained model performs worse than models trained separately on each task. The second issue is \textbf{Data Imbalance}. Significant differences in the data amount collected from various sources (e.g., a ratio of 50:1 between Retrieval and Classification tasks) can create inductive bias, severely impacting the model\'s performance across different tasks.

In this work, we examine a simplified yet representative multi-task joint training setting involving text retrieval and semantic textual similarity (STS) tasks. Our empirical analysis confirms that task conflict arises in the commonly used joint training approach, leading to poorer performance in jointly trained models compared to those trained separately on individual tasks. Furthermore, we utilize task vectors, as described by ~\cite{ilharco_editing_2023}, to analyze the impact of data imbalance on model performance. Task vectors assess the shift in model weights across tasks, helping us understand performance disparities. Our findings indicate that the amount of training data significantly influences the positioning of task vectors (see Figure~\ref{fig:task_conflits_and_data_imbalance}), thereby affecting model performance on different tasks.

To address the limitations of directly joint-training general text embedding models while ensuring their generalization and robustness, we explore model merging, a technique gaining popularity in multi-task learning scenarios. This method combines parameters from different models, allowing the unified model to handle multiple tasks without the gradient conflicts and data imbalance issues inherent in joint training. We evaluate five common model merging strategies in the context of general text embedding. The results show significant improvements in multi-task performance. We also analyze the relationship between merged models and the interpolation space of task vectors, focusing on how the direction and norm of merged models influence performance. Our findings indicate that \textbf{\textit{searching within the interpolation space of task vectors is sufficient to identify the optimal model merge strategy}}. 

In model merging method, each sub-model is trained independently, and the hyperparameters increase linearly with the number of sub-models, complicating parameter searches. We find that the performance of merged models correlates with their loss on multi-task training data. Therefore, we propose an efficient method called \textbf{Self Positioning} to search for optimal merging positions within the interpolation space of task vectors. This technique uses stochastic gradient descent to find low-loss hyperparameter combinations for a specified target dataset. Our experiments confirm that \textbf{Self Positioning} identifies high-performance merging results effectively. Compared to resampling methods for handling data imbalance, Self Positioning achieves better performance with significantly less time cost. To fully validate the effectiveness of Self Positioning method, we apply Self Positioning to a process involving 330 tasks collected by Instructor~\cite{su_one_2023}. Experiments demonstrate that our model merging strategy significantly improved performance on the MTEB dataset (0.7 point increasement in average performance).

Briefly, our main contributions are as follows: 1) We identified two key problems affecting the performance of general text embeddings in joint training: \textbf{Task Conflict} and \textbf{Data Imbalance}, through detailed experimentation and analysis. 2) We explore using model merging to address these limitations. Additionally, we discovered that high-performing merged models can be found within the interpolation space of task vectors. This inspired us to propose a \textbf{Self Positioning} approach as an automatic search method for model merging. 3) We conduct extensive experiments and analysis to validate the effectiveness of Self Positioning, significantly enhancing the performance of general text embedding models.

\section{Background}
\label{sec:background}

\subsection{General Text Embeddings}
\label{sec:background-general-text-embeddings}
Text embeddings have shown significant potential in applications such as web retrieval, text similarity, and text classification. Initially, research focused on creating task-specific embeddings. However, with the emergence of pretrained language models, the emphasis has shifted toward developing general text embedding models capable of handling various tasks. Studies like GTR~\cite{ni_large_2022}, E5~\cite{wang2022text}, GTE~\cite{li_towards_2023}, BGE~\cite{xiao_c-pack_2023}, and JinaEmbedding~\cite{gunther2023jina} have fine-tuned embedding models using large amounts of multi-task relevance data. Additionally, there are studies like TART~\cite{asai2023task} and InstructOR~\cite{su_one_2023} that introduce natural language prompts to guide embedding models in generating task-specific outputs. Leveraging the powerful text modeling capabilities of pretrained Large Language Models (LLMs), models like SGPT~\cite{sgpt}, UDVER~\cite{Zhang2023LanguageMA}, E5Mistral~\cite{wang_improving_2024}, RepLLaMA~\cite{ma2024fine}, GTE-Qwen~\cite{li_towards_2023}, and NV-Embed~\cite{lee2024nv} utilize LLMs as their foundations, refining them with multi-task data and instructional guidance. 

To enhance the performance of general text embedding models, recent research primarily focuses on optimizing training data and pretrained language models. In terms of training data, efforts concentrate on collecting large-scale, high-quality, multi-domain, and multi-task relevant text pairs from public benchmarks or web resources~\cite{ni_large_2022,wang2022text,li_towards_2023,xiao_c-pack_2023}. Synthetic data is also generated using large language models to expand the dataset further~\cite{wang_improving_2024}. For pretrained models, researchers aim to adapt models designed for general natural language understanding (NLU) or generation tasks to better suit embedding tasks~\cite{xiao-etal-2022-retromae,li-etal-2024-llama2vec}. During model training, most approaches employ contrastive learning as the optimization objective~\cite{oord_representation_2018}. Consider the training data $\mathcal{D}$ consists of $N$ datasets: $\mathcal{D} = \{D_i\}_{i=1}^N$, each containing $N_i$ instances: $D_i = \{I_j\}_{j=1}^{N_i}$. Each instance $I_j$ includes a query text $q$, a similar text $p$, and a set of dissimilar texts $P^\mathrm{n} = \{p^\mathrm{n}_k\}_{k=1}^K$. A model $M$ with parameters $\theta$ maps each text to an L2-normalized vector $M(p)$. For each instance $I_j$, we optimize the following loss function:
\begin{equation}
    \mathcal{L}^\mathrm{CL}_{I_j} \notag
  = -\log\frac{\exp(M(p)^\top M(q)/\tau)}{\exp(M(p)^\top M(q)/\tau) + \sum_{k=1}^K \exp(M(p)^\top M(q^\mathrm{n}_k)/\tau)},
\end{equation}
where $\tau$ is a temperature parameter.

Unlike previous studies, this work focuses on the training process of models, particularly identifying and addressing the limitations of joint training with multi-task large-scale datasets to enhance the performance of the final model.

\subsection{Task Vectors}
\label{sec:background_task_vectors}

The concept of task vectors was introduced by~\cite{ilharco_editing_2023}. Given an initial model $M_0$ with weights $\theta_0$, and a trained model $M_i$ with weights $\theta_i$ after learning on a specific task $i$, the task vector for the corresponding task is defined as:
\begin{equation}
  V_i = \theta_i - \theta_0.
\end{equation}
\cite{ilharco_editing_2023} discovered that performing arithmetic operations on task vectors directly affects the model's performance on the corresponding tasks. For instance, adding or subtracting the task vector $V_i$ to the weights $\theta_j$ of another model $M_j$ will respectively increase or decrease $M_j$'s performance on task $i$. Therefore, task vectors effectively represent the capabilities a model has acquired for specific tasks.

\begin{figure}
    \centering
    \subfloat[The difference in model performance obtained from joint training compared to the best performance achieved when trained separately on STS and Retrieval tasks.]{
        \includegraphics[width=.36\linewidth]{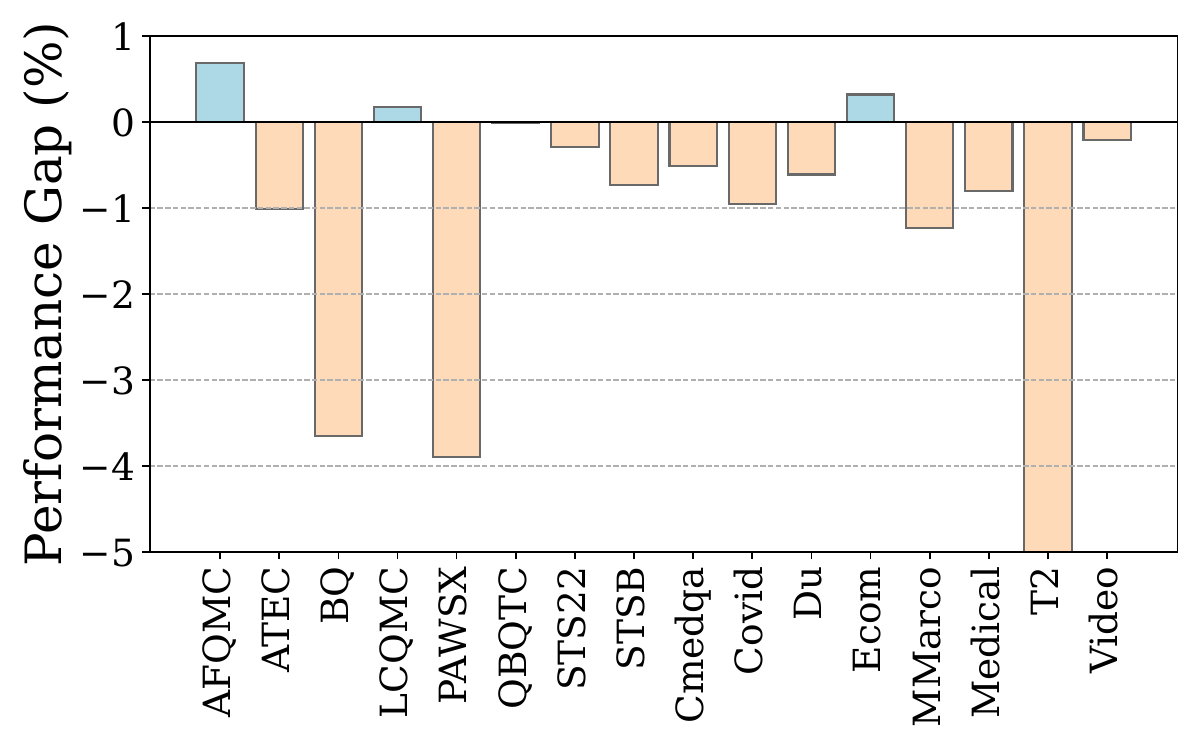}
        \label{fig:problems_task_conflicts}
    }
    \hfill
    \subfloat[The Impact of Data Amount on Norm]{
        \includegraphics[width=.18\linewidth]{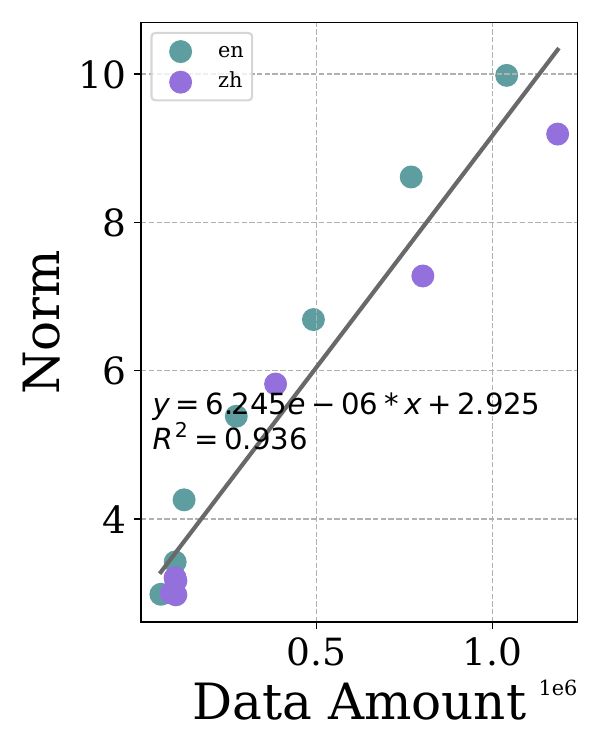}
        \label{fig:data_imbalance_impact_on_norm}
    }
    \hfill
    \subfloat[Subjects of statistics]{
        \includegraphics[width=.17\linewidth]{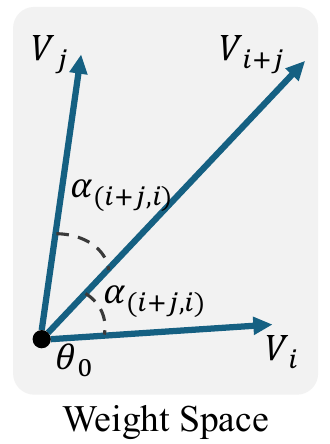}
        \label{fig:analysis_direction_illustration}
    }
    \hfill
    \subfloat[The Impact of Data Amount on Direction]{
        \includegraphics[width=.18\linewidth]{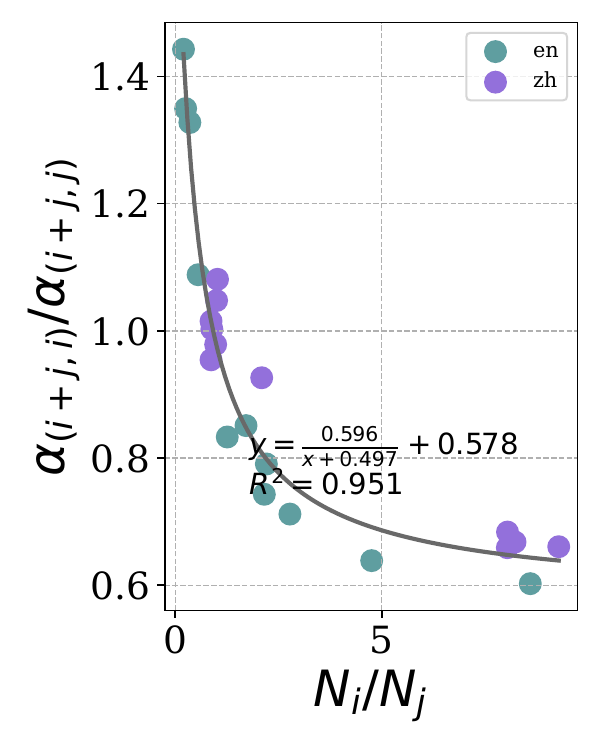}
        \label{fig:data_imbalance_impact_on_direction}
    }
    \caption{Experimental Results on Task Conflict and Data Imbalance Problems.}
    \label{fig:task_conflits_and_data_imbalance}
\end{figure}

This study investigates how the norm and direction of task vectors impact task performance. The norm of the task vector, $\|V_i\|$, measures its length in the weight space. The direction of the task vector pertains to the relative orientation between the combined task vector $V_{i+j}$, which can perform tasks $i$ and $j$ simultaneously, and the individual task vectors $V_i$ and $V_j$, which can perform tasks $i$ and $j$ independently. Specifically, we define direction as:
\begin{equation}
\label{eq:direction}
    \frac{\alpha_{(i+j,i)}}{\alpha_{(i+j,i)} + \alpha_{(i+j,j)}}, 
\end{equation}
where $\alpha_{(i+j,i)}$ is the angle between $V_{i+j}$ and $V_i$, and $\alpha_{(i+j,j)}$ is the angle between $V_{i+j}$ and $V_j$ (see Figure~\ref{fig:analysis_direction_illustration} for illustration). Equation~\ref{eq:direction} quantifies the extent to which the task vector $V_{i+j}$ aligns more closely with $V_i$ relative to $V_j$.

\subsection{Model Merging}
\label{sec:background_model_merging}
Model merging~\cite{wortsman_model_2022,xiao_lm-cocktail_2024,yu_language_2024,yang_model_2024} refers to the process of integrating the parameters of multiple models with the same architecture into a single unified model, enabling the merged model to possess the capabilities of multiple models. Model merging has been applied to many machine learning subfields, such as continual learning to mitigate catastrophic forgetting of old tasks, and multi-task/multi-domain learning to facilitate knowledge transfer~\cite{yang_model_2024}.

This work involves five popular model merging methods. Specifically, given $N$ datasets $\{D_i\}_{i=1}^N$ and the $N$ models $\{M_i\}_{i=1}^N$ trained on them, with corresponding model weights $\{\theta_i\}_{i=1}^N$ and task vectors $\{V_i\}_{i=1}^N$, the following five methods are used to obtain the merged task vector $V_\mathrm{m}$: 
(1) \textbf{Average} Merging~\cite{wortsman_model_2022} performs linear interpolation between different task vectors to obtain $V_\mathrm{m}$:
\begin{equation} 
    V_\mathrm{m} = \sum_{i=1}^N w_i V_i\left/\sum_{i=1}^N w_i\right.,
\end{equation}
where $\{w_i\}_{i=1}^N$ ($w_i\in\mathbb{R}$) are hyperparameters; 
(2) \textbf{SLERP}~\footnote{\url{https://github.com/Digitous/LLM-SLERP-Merge}} performs \textbf{S}pherical \textbf{L}inear int\textbf{ERP}olation successively between two task vectors for $(N-1)$ times to obtain $V_\mathrm{m}$: 
\begin{align}\label{eq:slerp}
    V_{i+j} 
    &= \frac{\sin\left(\frac{a_j}{a_i + a_j}\alpha_{(i,j)}\right)V_i + \sin\left(\frac{a_i}{a_i + a_j}\alpha_{(i,j)}\right)V_j}{\sin(\alpha_{(i,j)})},
\end{align}
where $a_i,a_i\in\mathbb{R}$ are hyperparameters; 
(3) \textbf{TIES} Merging tackles the task conflicts in task vectors by trimming low-magnitude model weights, resolving sign disagreements, and disjointly merging parameters with
consistent signs to obtain $V^\mathrm{TIES}$, which is scaled by hyperparameter $\lambda$ to obtain $V_\mathrm{m}$:
\begin{equation} 
    V_\mathrm{m} = \lambda V^\mathrm{TIES}; 
\end{equation}
(4) \textbf{Fisher} Merging~\cite{matena_merging_2022} calculates merging coefficient $\hat{F}_{\theta_i} \in \mathbb{R}^{\|\theta_0\|}$ based on Fisher information matrix: 
\begin{equation} 
    \hat{F}_{\theta_i} = \frac{1}{N_i}\sum_{j=1}^{N_i} (\nabla_{\theta_i}\mathcal{L}^\mathrm{CL}(\theta_{i};I_{j}))^2, 
\end{equation} 
where $I_{j} \in D_i$ is a training sample. $\hat{F}_{\theta_i}$ is then used to obtain $V\mathrm{m}$: 
\begin{equation} 
    V_\mathrm{m}^{(k)} = \frac{\sum_{i=1}^N \lambda_i F_{\theta_i}^{(k)} V_i^{(k)}}{\sum_{i=1}^N \lambda_i F_{\theta_i}^{(k)}}, 
\end{equation}
where $\{\lambda_i\}_{i=1}^N$ ($\lambda_i\in\mathbb{R}$) are hyperparameters, and $k=1,\cdots,\|\theta_0\|$; 
(5) \textbf{RegMean} Merging~\cite{jin_dataless_2023} performs model merging by optimizing a linear regression problem with the following closed-form solutions:
\begin{equation} 
    W_\mathrm{m}^k = (\sum_{i=1}^N X_i^k(X_i^k)^\top)^{-1} \sum_{i=1}^N (X_i^k (X_i^k)^\top W_i^k)
\end{equation}
to obtain $V_\mathrm{m}$, where $\{W_i^k\}_{k=1}^K$ are linear layers weights in $\theta_i$, each corresponding to an input matrix $\{X_i^k\}_{k=1}^K$. We implement these methods based on the implementations from \cite{yu_language_2024}.

\section{Problems Analysis}
\label{sec:problems}

\subsection{Training and Evaluation Setup}
\label{sec:problems_setup}

To investigate the problem of {\bf task conflict} and {\bf data imbalance} in the joint-training process of embedding models, we first conduct experiments in a controlled setting. Specifically, we utilize ten datasets for training across two tasks: Semantic Textual Similarity (STS) and Retrieval, in both English and Chinese languages. These two tasks were selected due to their widespread application in real-world scenarios and their differing data requirements—STS involves matching symmetric texts, while Retrieval often requires matching asymmetric texts. This selection naturally highlights potential conflicts between the tasks.

\paragraph{Dataset} For the English STS task, we use the AllNLI~\cite{gao_simcse_2021} dataset, and the Retrieval task incorporates four datasets: FEVER~\cite{thorne_fever_2018}, HotpotQA~\cite{yang_hotpotqa_2018}, MS MARCO~\cite{nguyen_ms_2016}, and NQ~\cite{kwiatkowski_natural_2019}.
In the Chinese STS task, we utilize the SimCLUE dataset\footnote{\url{https://github.com/CLUEbenchmark/SimCLUE}}. The Retrieval task for Chinese includes DuReaderRetrieval~\cite{qiu_dureader-retrieval_2022}, Ecom/Medical/Video-Retrieval~\cite{long_multi-cpr_2022}. The size of each dataset is detailed in Table~\ref{tab:data_amount_sts_retrieval}.

\paragraph{Training Details} We train the $\mathrm{BERT_{base}}$ model~\cite{devlin_bert_2019} separately for English and Chinese using the aforementioned datasets. Following the training approach proposed by  \cite{li_towards_2023}, each batch contains data from a single task. The hyperparameters for all experiments are set as follows: 3 epochs, a batch size of 256, a learning rate of $1 \times 10^{-5}$, a temperature $\tau$ of $2 \times 10^{-2}$, and a warmup ratio of 0.1.

\paragraph{Evaluation} The model's performance is evaluated using the STS and Retrieval evaluation tasks from MTEB~\cite{muennighoff_mteb_2023}, which include 25 tasks in English and 16 tasks in Chinese.

\begin{table}
  \centering
  \small
  \caption{The number of instances in training datasets.}
  \begin{tabular}{c|c|cccc}
    \toprule
          & \textbf{STS}   & \multicolumn{4}{c}{\textbf{Retrieval}} \\
    \midrule
    \multirow{2}[2]{*}{\textbf{en}} & ALLNLI & FEVER & HotpotQA & MS MARCO & NQ \\
          & 271,612 & 123,140 & 97,825 & 491,007 & 57,110 \\
    \midrule
    \multirow{2}[2]{*}{\textbf{zh}} & SimCLUE & DuReader & Ecom  & Medical & Video \\
          & 802,291 & 86,395 & 99,763 & 97,509 & 99,719 \\
    \bottomrule
  \end{tabular}%
  \label{tab:data_amount_sts_retrieval}%
\end{table}%

\subsection{Task Conflict}
\label{sec:problems_task_conflicts}

Research on multi-task learning has shown that jointly training multiple tasks can lead to the \textit{Negative Transfer Phenomenon}~\cite{zhang_survey_2022}, where the model's performance deteriorates compared to models trained separately on each task. We observe a similar effect in embedding model training. Specifically, we trained models separately on English and Chinese STS and Retrieval datasets and then evaluated their performance on each task, as illustrated in Figure~\ref{fig:problems_task_conflicts}. The results demonstrate that, in the Chinese setting, the jointly trained model underperforms in 13 out of 16 evaluation tasks compared to task-specific models. Similarly, in the English setting, performance declines in 12 of 25 tasks (see Figure~\ref{fig:problems_task_conflicts_en} in the appendix). Previous studies~\cite{yu_gradient_2020} suggest that this may be due to detrimental gradient interference between tasks, where optimizing for one task conflicts with the gradients of others. This observation motivates the solutions we propose in Section~\ref{sec:solutions}.

\subsection{Data Imbalance}
\label{sec:problems_data_imbalance}

As shown in Table~\ref{tab:data_amount_sts_retrieval}, the sizes of different datasets can vary by nearly a factor of 10. However, current methods do not account for this imbalance and instead, combine all datasets in their entirety for training.
To thoroughly understand the issue of data imbalance, we examine how differences in training data size affect the performance of embedding models in this section. 

\subsubsection{Impact on Norm}

First, we explore the relationship between the amount of training data and the norm of the task vector in the trained model. We train the model on individual datasets (e.g., DuReaderRetrieval training data), on single-task datasets (e.g., all Retrieval training data), and on all training datasets combined. We then analyze the relationship between the norm of the task vectors ($\|V_i\|$) and the amount of training data ($N_i$). The results, shown in Figure~\ref{fig:data_imbalance_impact_on_norm}, reveal a clear linear relationship, indicating that the norm of the task vectors increases proportionally with the amount of training data.

\subsubsection{Impact on Direction}
Next, we investigate the relationship between the amount of training data and the direction of the task vectors. For a pair of training datasets $(D_i, D_j)$ with respective data sizes $N_i$ and $N_j$, we perform the following steps: 1) Train jointly on both datasets to obtain model $M_{i+j}$ and its task vector $V_{i+j}$. 2) Train separately on each dataset to obtain models $M_i$ and $M_j$, with task vectors $V_i$ and $V_j$, respectively. 3) Measure the angles $\alpha_{(i+j,i)}$ between $V_{i+j}$ and $V_i$, and $\alpha_{(i+j,j)}$ between $V_{i+j}$ and $V_j$. Figure~\ref{fig:analysis_direction_illustration} illustrates the experimental setup. The dataset pairs $(D_i, D_j)$ include individual pairs (e.g., SimCLUE and DuReaderRetrieval) as well as a combination of all STS and Retrieval training data. To clearly demonstrate the relationship between the angle ratios $\frac{\alpha_{(i+j,i)}}{\alpha_{(i+j,j)}}$ and the data size ratios $\frac{N_i}{N_j}$, we plot these values in Figure~\ref{fig:data_imbalance_impact_on_direction}. The figure shows that as $\frac{N_i}{N_j}$ increases, the ratio $\frac{\alpha_{(i+j,i)}}{\alpha_{(i+j,j)}}$ decreases significantly. This indicates that the task vector obtained from joint training becomes more aligned with the task vector of the dataset that has a larger amount of training data.

These experiments reveal that differences in dataset sizes introduce varying inductive biases, significantly affecting both the norm and direction of the task vectors. This suggests that the properties of the task vectors are heavily influenced by the amount of training data, serving as an inductive bias that does not directly correlate with the model's performance. Consequently, it is possible that task vectors with better performance exist at different positions.

\section{Solutions via Model Merging}
\label{sec:solutions}

According to the experiments in Section~\ref{sec:problems}, joint training presents two main challenges: 1) Negative Transfer: Gradient conflicts between different tasks during optimization; 2) Data Imbalance: The amount of training data significantly affects the positioning of task vectors.
We propose that model merging can address these issues effectively because: 1) Independent Training: Model merging ensures that different tasks are trained independently, thereby avoiding gradient conflicts during optimization. 2) Adjustable Task Vectors: Task vectors can be scaled and weighted during the merging process, which helps mitigate or eliminate the effects of data imbalance.
In this section, we investigate the effectiveness of model merging in training embeddings.

\subsection{Model Merging Pipelines}
\label{sec:solutions_mergin_pipelines}

Based on model merging, we come up with the two training pipelines, both of which have potential applications in training embedding models: (1) \textbf{Separate Merging}: Using distinct datasets $\{D_i\}_{i=1}^N$, we train separate models $\{M_i\}_{i=1}^N$ from a common backbone. These models are then merged to form the final model $M_\mathrm{m}^\mathrm{sep}$. This approach enables the transformation of an existing backbone (e.g., a language model) into an embedding model. (2) \textbf{Iterative Merging}: We iteratively train the backbone on the datasets $(D_1, \dots, D_N)$ to obtain a sequence of models $(M_1, M_{(1,2)}, \dots, M_{(1,\dots,N)})$. These models are subsequently merged to create the final model $M_\mathrm{m}^\mathrm{iter}$. This method is particularly effective for enhancing the performance of existing embedding models on specific tasks.

\subsection{Effectiveness of Model Merging}
\label{sec:solutions_effectiveness}

\begin{figure}[ht]
    \centering
    \includegraphics[width=0.80\linewidth]{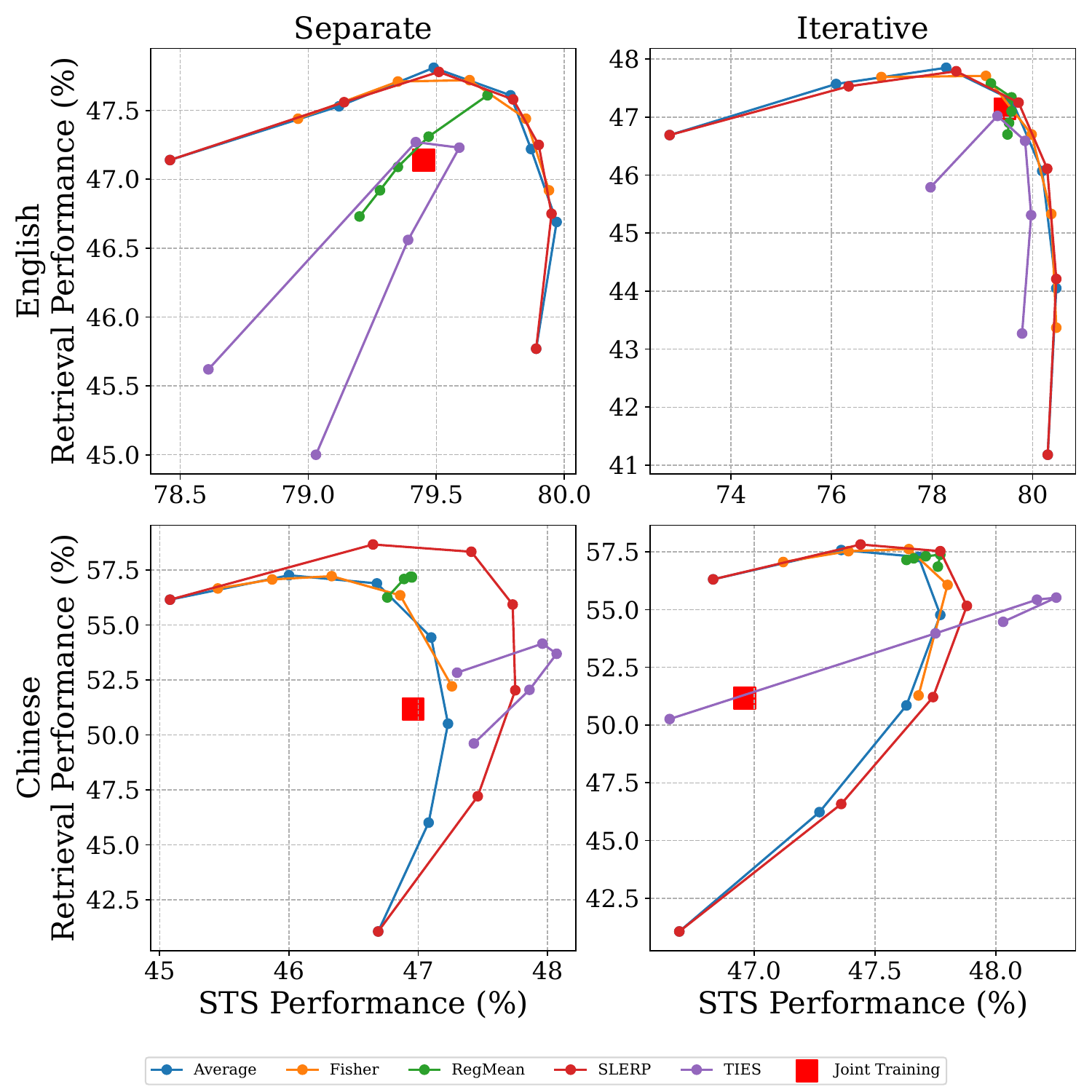}
    \caption{Performance of models under different languages, merging settings, and model merging methods during hyperparameter tuning.}
    \label{fig:hp_tuning}
\end{figure}

To validate the training pipelines described in Section~\ref{sec:solutions_mergin_pipelines}, we evaluate them using the model merging methods outlined in Section~\ref{sec:background_model_merging} within the simplified settings proposed in Section~\ref{sec:problems}. Specifically, we use training data from STS and Retrieval to obtain the task vectors $V_\mathrm{s}$ and $V_\mathrm{r}$ for their respective tasks and then derive the merged task vector $V_\mathrm{m}$ through various model merging techniques. The training configurations align with those detailed in Section~\ref{sec:problems_setup}. For each model merging method, we optimize performance through hyperparameter tuning, with results presented in Figure~\ref{fig:hp_tuning} and Table~\ref{tab:performance_sts_retrieval}. Additionally, we include results from joint training in the figure. In the figure, models positioned closer to the top right corner demonstrate stronger average multi-task performance. Our comparison reveals that, across both English and Chinese settings and the two training pipelines, model merging methods significantly outperform joint training in terms of average performance.

The improvement in average performance from model merging demonstrates its effectiveness in the application and highlights its capability to alleviate task conflict during training.

\subsection{Key Factors in Model Merging}
\label{sec:solutions_key_factors}
In this section, we explore the key factors influencing the merged performance. From the results of Section~\ref{sec:solutions_effectiveness}, it is evident that the performance improvements brought by different model merging methods vary, with SLERP performing well in all settings and having the highest upper-performance limit in three out of four settings. Therefore, we mainly study the SLERP method and use the Averaging method, 
whose merging results both lie in the interpolation space of task vectors, for comparison.

\subsubsection{Direction of $V_\mathrm{m}$}

\begin{figure}[ht]
    \centering
    \includegraphics[width=.9\columnwidth]{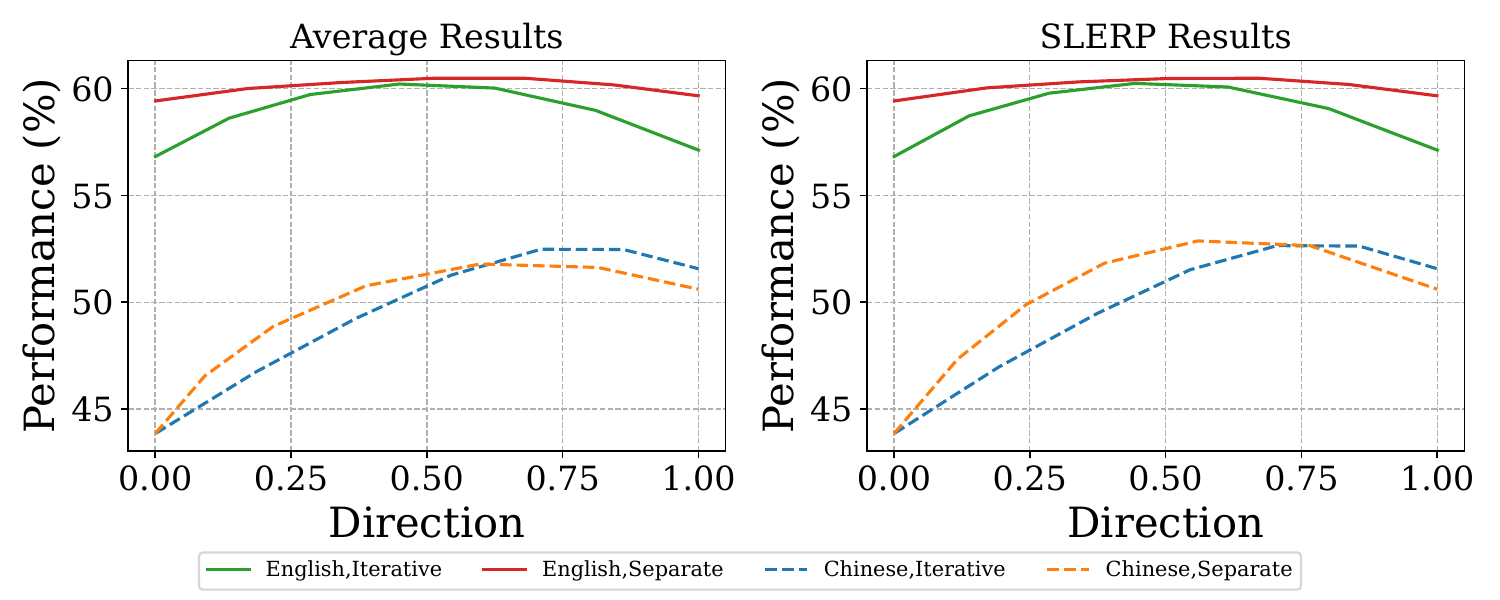}
    \caption{The relationship between the direction of task vectors and the average performance on STS and Retrieval tasks in different settings.}
    \label{fig:factors_direction}
\end{figure}

\begin{figure}[h]
    \centering
    \includegraphics[width=.8\columnwidth]{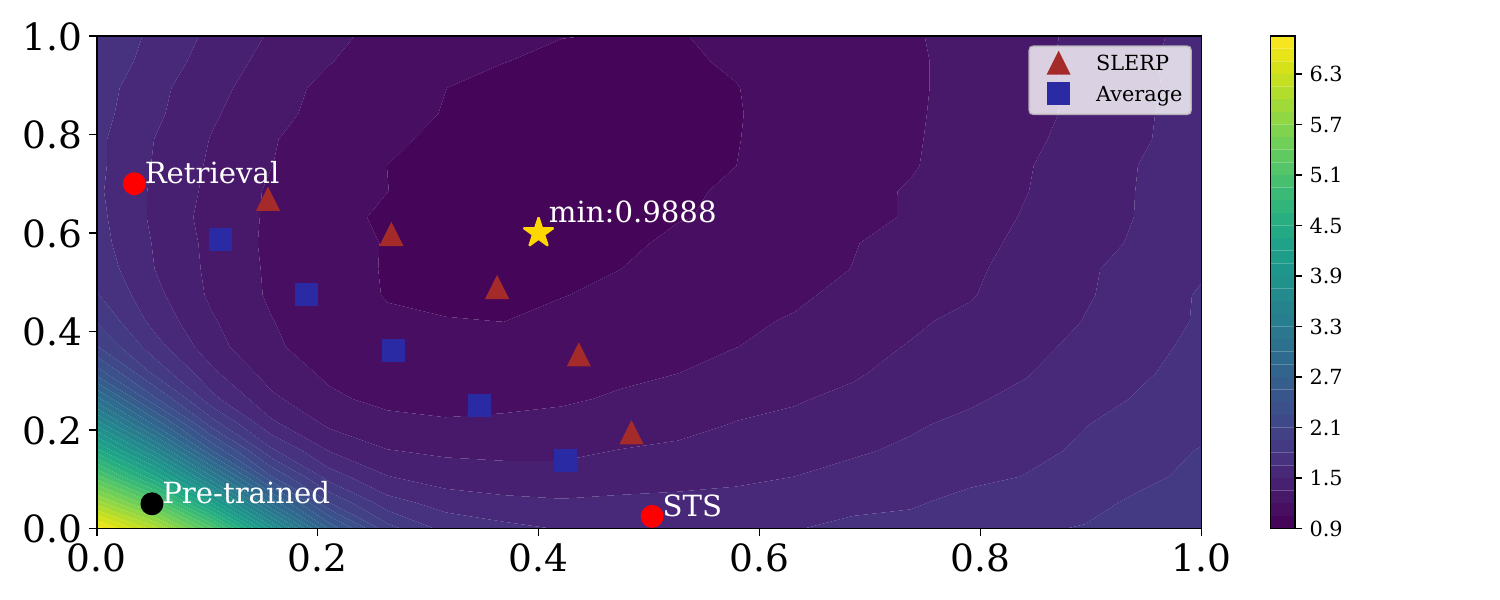}
    \caption{The loss landscape on the interpolation space of Retrieval and STS task vectors, under English Separate Merging. Other settings yield similar results.}
    \label{fig:factors_norm}
\end{figure}

We examine the link between model performance and the direction of task vectors using the Average and SLERP methods. The results are plotted in Figure~\ref{fig:factors_direction}, where we can observe that direction significantly impacts the performance of the task vector. Therefore, \textit{direction is a key factor when seeking the optimal task vector}.

\subsubsection{Norm of $V_\mathrm{m}$}

Though both Average and SLERP can explore many directions, there is a performance gap between them (see Figure~\ref{fig:hp_tuning}). To investigate this gap, we plot the loss landscape under multi-task training data on the interpolation space $P_{(s,r)}$ in Figure~\ref{fig:factors_norm} (refer to Appendix~\ref{sec:loss_landscape} for statistical details), and plot the merging results $V_\mathrm{m}$ from Average and SLERP methods in it. From the figure, we can see: 1) The norm of $V_\mathrm{m}$ from SLERP is consistently longer than that from Average; 2) $V_\mathrm{m}$ from SLERP consistently stays closer to regions with lower loss in the loss landscape. These findings reflect the impact of the norm of $V_\mathrm{m}$ on its properties (i.e., the loss on multi-task training data).  
Due to the differences in norm between the Average and SLERP methods leading to variations in performance, this indicates that \textit{norm is a key factor when seeking the optimal task vector}.

\section{Self Positioning in the Interpolation Space}
\label{sec:self_positioning}

As shown in Figure \ref{fig:hp_tuning}, the performance upper bound (The point closest to the top right corner) is sensitive to both the direction and norm of the task vector, which is only reached under a few merging weight settings. When the number of models is small (e.g., $N=2$), different combinations of directions and norms can be explored using a grid search method to achieve high performance. However, as the number of parameters describing directions increases linearly with the number of models $N$ (referring to the SLERP merging process), the time complexity of the grid search exponentially increases with the number of models $N$. This makes the method of finding the merging result $V\mathrm{m}$ through exhaustive search impractical. Therefore, in this section, we propose a method to find a near-optimal $V_\mathrm{m}$ within the interpolation space. 

\subsection{Methodology}
\label{sec:self_positioning_methodology}

Our approach is essentially a hyperparameter search method based on a probe dataset, therefore, it can be applied to any model merging method. Given SLERP's superior performance, we use the SLERP method for all equations and experiments. For brevity, we rewrite Equation \ref{eq:slerp} in functional form:

\begin{equation}
    f^\mathrm{SLERP}(V_i, V_j, a_i, a_j) \notag
    = \frac{\sin\left(\frac{a_j}{a_i + a_j}\alpha_{(i,j)}\right)V_i + \sin\left(\frac{a_i}{a_i + a_j}\alpha_{(i,j)}\right)V_j}{\sin(\alpha_{(i,j)})},
\end{equation}

Given $N$ task vectors $\{V_i\}_{i=1}^N$ and a specific model merging method, we try to control the direction and norm of the merged task vector using different learnable parameters.

\paragraph{Direction Control} Corresponding merging weights $\{a_i\}_{i=1}^N$, we merge them in an iterative form:
\begin{equation}
  V_{(1,\cdots,N)} = f^\mathrm{SLERP}\left(V_{(1,\cdots,N-1)}, V_N, \frac{\sum_{i=1}^{N-1} a_i}{N-1}, a_N\right), 
\end{equation}

\paragraph{Norm Control} By applying a scaling factor $\lambda$, we can 
represents all positions in the interpolation space:
\begin{equation}
  V_\mathrm{p} = \lambda V_{(1,\cdots,N)}
\end{equation}
As our goal is to search for positions with low loss, we need to construct a target dataset $D_\mathrm{t}$, and form the optimization problem:
\begin{align}\label{eq:optimization_objective}
  (\{\hat{a_i}\}_{i=1}^N, \hat{\lambda}) = \argmin_{(\{a_i\}_{i=1}^N, \lambda)}\left(\frac{1}{\left|D_\mathrm{t}\right|}\sum_{I\in D_\mathrm{t}} \mathcal{L}^\mathrm{CL}(I; \theta_0 + V_\mathrm{p}) + \mu \lambda\right)
\end{align}
where $\theta_0$ represents the init parameters of the embedding model, and $\mu$ is a hyperparameter used to prevent overfitting in this optimization problem. For the target dataset $D_\mathrm{t}$ being used, we note that: 1) Since the number of parameters to be optimized is small, the amount of data required for the target dataset is much less than that required to train the model; 2) The target dataset can be obtained either by sampling from the training dataset or by choosing to construct it from other data sources. Overall, we can find the final task vector by solving this optimization problem using stochastic gradient descent algorithm~\cite{andrychowicz_learning_2016}. We refer to this method of finding the merging results within the interpolation space of task vectors as \textbf{Self Positioning}.

\subsection{Experiments}
\label{sec:self_positioning_experiments}

\begin{table}
  \centering
  \caption{Average performance (\%) on all STS and Retrieval tasks. ``JT'' represents results from joint training, ``SP'' represents results from Self Positioning, and other model merging methods report the best results after hyperparameter tuning.}
  \small
    \begin{tabular}{c|c|c|ccccc|c}
    \toprule
          & \makecell{\textbf{JT}} & \makecell{\textbf{Merging}\\ \textbf{Pipeline}} & \makecell{\textbf{Fi-}\\\textbf{sher}} & \makecell{\textbf{Reg}\\ \textbf{Mean}} & \textbf{TIES}  & \makecell{\textbf{Ave-}\\\textbf{rage}} & \makecell{\textbf{SL-}\\\textbf{ERP}} & \makecell{\textbf{SP}} \\
    \midrule
    \multirow{2}[2]{*}{\textbf{en}} & \multirow{2}[2]{*}{60.1} & Separate   & \textbf{60.5} & 60.4  & 60.2  & \textbf{60.5} & \textbf{60.5} & \textbf{60.5} \\
          &       & Iterative  & \textbf{60.3}    & 60.2 & 59.9  & 60.2 & 60.2 & 60.1 \\
    \midrule
    \multirow{2}[2]{*}{\textbf{zh}} & \multirow{2}[2]{*}{51.2} & Separate   & 51.7  & 52.1  & 51    & 51.8  & \textbf{52.9} & \textbf{52.9} \\
          &       & Iterative  & 52.6  & 52.6  & 51.9  & 52.5  & 52.6 & \textbf{52.9} \\
    \bottomrule
    \end{tabular}%
  \label{tab:performance_sts_retrieval}%
\end{table}

We apply Self Positioning in the model merging settings described in Section~\ref{sec:solutions_effectiveness}, which includes four merging processes: separate merging and sequential merging for both English and Chinese. To construct the target dataset $D_\mathrm{t}$, we sample data from the training datasets for the STS task and the Retrieval task in equal proportions, resulting in a final $D_\mathrm{t}$ that contains $32,000$ instances. The optimization process uses the following settings: the optimizer is \verb|Adam|~\cite{kingma_adam_2015}, the batch size is $32$, the learning rate is $5 \times 10^{-3}$, and the number of training steps is $1,000$. The initial values for the weight parameters $\{a_i\}_{i=1}^N$ and the scaling factor $\lambda$ are both set to $1$. We conduct experiments under $\mu \in \{0.00, 0.05, 0.10\}$, with the best results presented in Table~\ref{tab:performance_sts_retrieval}. The results in the table show that Self Positioning achieved the best results in three of the four merging settings. Although it does not achieve the best results in the iterative merging for English, it performs comparably to joint training. It is important to note that the results of other model merging methods in the table are the best results obtained after hyperparameter tuning. Therefore, the performance comparison in the table demonstrates the effectiveness of Self Positioning in searching for merging results with good performance.

\begin{figure}[t]
    \centering
    \subfloat[Results when fixing $\lambda$]{
        \includegraphics[width=.48\columnwidth]{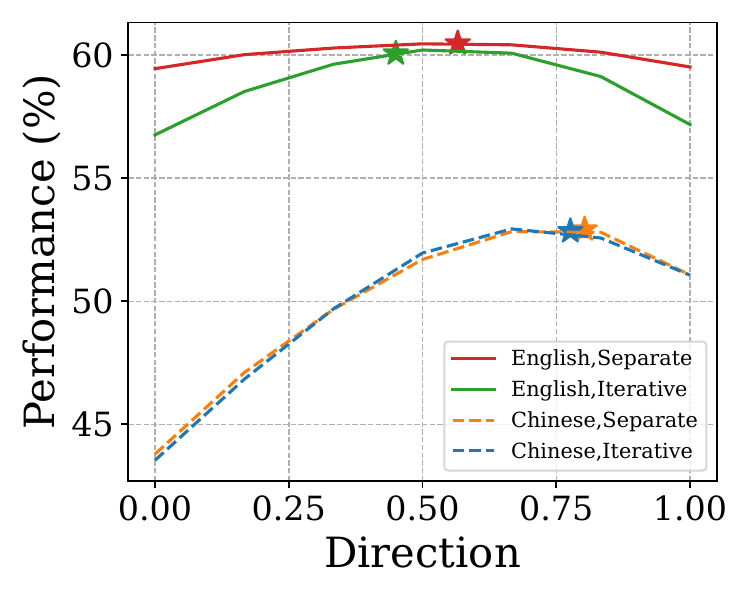}
        \label{fig:plot_self_positioning_effectiveness_fix_lambda}
    }
    \hfill
    \subfloat[Results when fixing direction]{
        \includegraphics[width=.48\columnwidth]{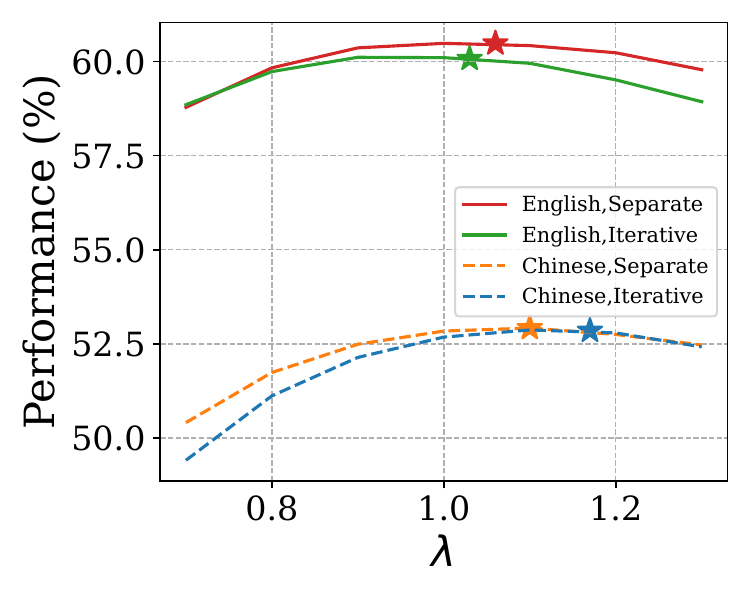}
        \label{fig:plot_self_positioning_effectiveness_fix_direction}
    }
    \caption{Experimental results for validating the effectiveness of the search results of Self Positioning.}
\end{figure}

\subsection{Performance Analysis}
\label{sec:self_positioning_analyses}

This section provides analyses of the merging results obtained through Self Positioning.

\subsubsection{Effectiveness of the Search Results}
\label{sec:self_positioning_analyses_effectiveness}
To determine whether the search results are positioned in areas of high performance within the interpolation space, we examine the performance surrounding these results. Specifically, for a search result $(\{\hat{a}_1, \hat{a}_2\}, \hat{\lambda})$, we conducted two experiments: (1) Holding \(\hat{\lambda}\) constant, we adjuste the ratio \(\frac{a_1}{a_1 + a_2}\) to take values in \(\left\{ \frac{0}{6}, \frac{1}{6}, \frac{2}{6}, \frac{3}{6}, \frac{4}{6}, \frac{5}{6}, \frac{6}{6} \right\}\) and evaluate the performance of the merging results. (2)  Keeping \(\{\hat{a}_1, \hat{a}_2\}\) fixed, we vary \(\lambda\) across the values \(\{0.7, 0.8, 0.9, 1.0, 1.1, 1.2, 1.3\}\) and assessed the performance of the merging outcomes. The results of Experiment (1) \& (2) are illustrated in Figure~\ref{fig:plot_self_positioning_effectiveness_fix_lambda} and Figure~\ref{fig:plot_self_positioning_effectiveness_fix_direction}, respectively. Additionally, the results identified by Self Positioning $(\{\hat{a}_1, \hat{a}_2\}, \hat{\lambda})$ are highlighted. Our findings indicate that the proposed method successfully identifies optimal results in most scenarios. In instances such as the English Separate Merging setting, where the optimal result was not achieved, the merging result remained closely aligned with the optimal position. This proximity underscores the effectiveness of the search results derived from Self Positioning.

\subsubsection{The Influence of $\mu$}
\label{sec:self_positioning_analyses_influence}

\begin{figure}[ht]
    \centering
    \includegraphics[width=.8\columnwidth]{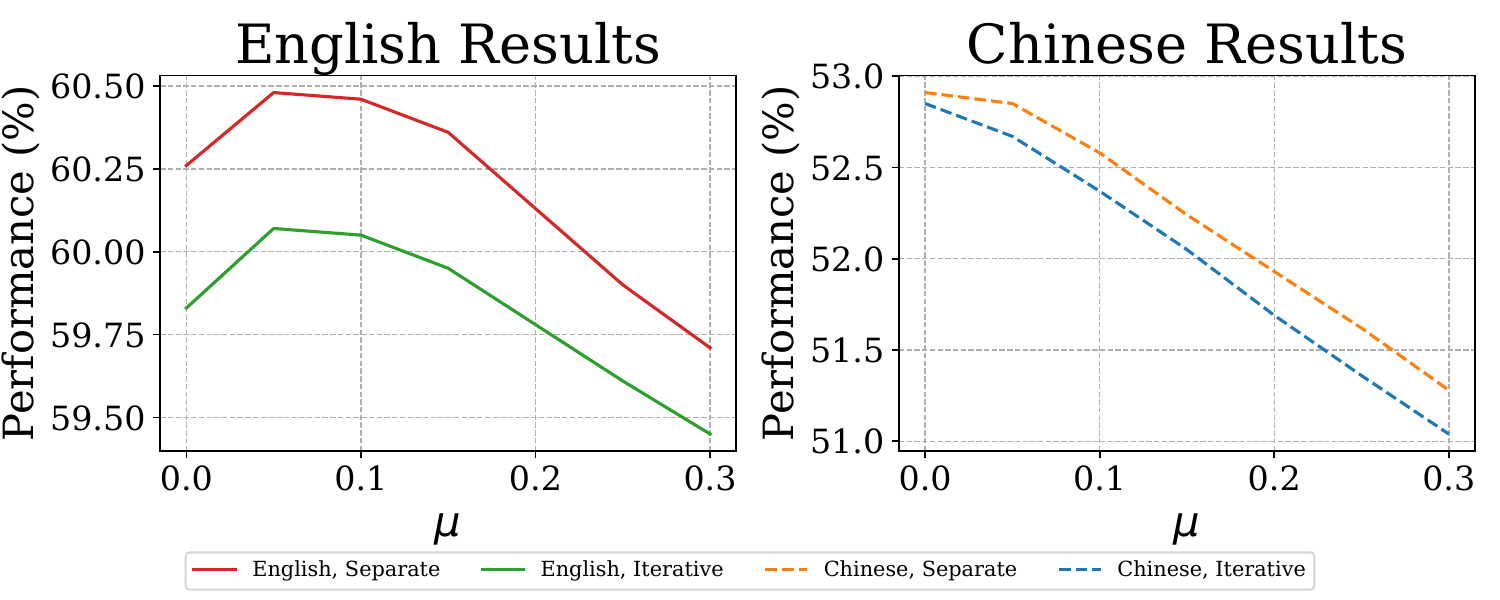}
    \caption{ The impact of the hyperparameter $\mu$ on the performance of Self Positioning search results.}
    \label{fig:self_positioning_ablation_mu}
\end{figure}

Equation~\ref{eq:optimization_objective} aims to prevent overfitting during optimization by constraining the scaling factor $\lambda$. In this section, we maintain the same settings as described in Section~\ref{sec:self_positioning_experiments} and conduct experiments with $\mu \in \{0.00, 0.05, 0.10, 0.15, 0.20, 0.25, 0.30\}$. Figure~\ref{fig:self_positioning_ablation_mu} visualizes the changes in the Self Positioning search results and model performance. Firstly, the changes in the search result positions indicate that $\mu$ significantly affects the magnitude but only slightly influences the direction. Secondly, the model performance results show that optimal performance is typically achieved when $\mu$ is set to a relatively low value (i.e., $\mu \leq 0.05$).

\begin{figure}[t]
    \centering
    \subfloat[Performance Comparison]{
        \includegraphics[width=.8\columnwidth]{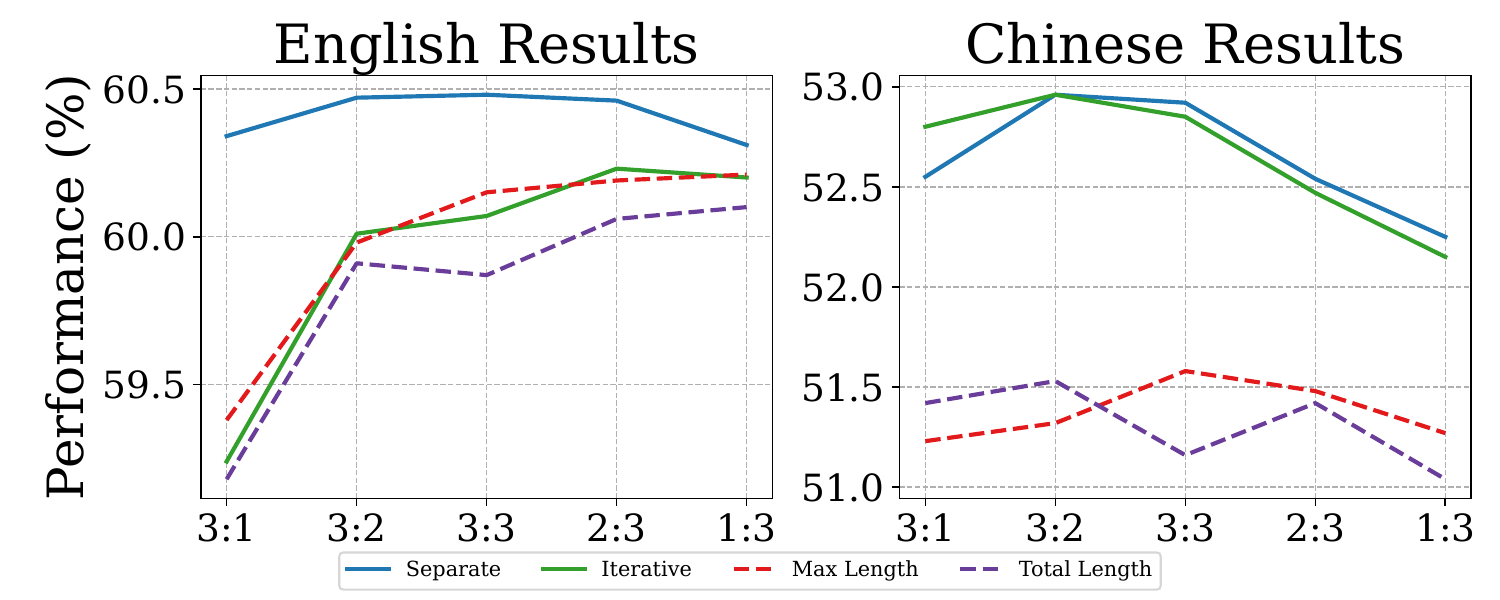}
        \label{fig:tack_data_imbalance_performance}
    }
    \\
    \subfloat[Time Cost Comparison]{
        \includegraphics[width=.8\columnwidth]{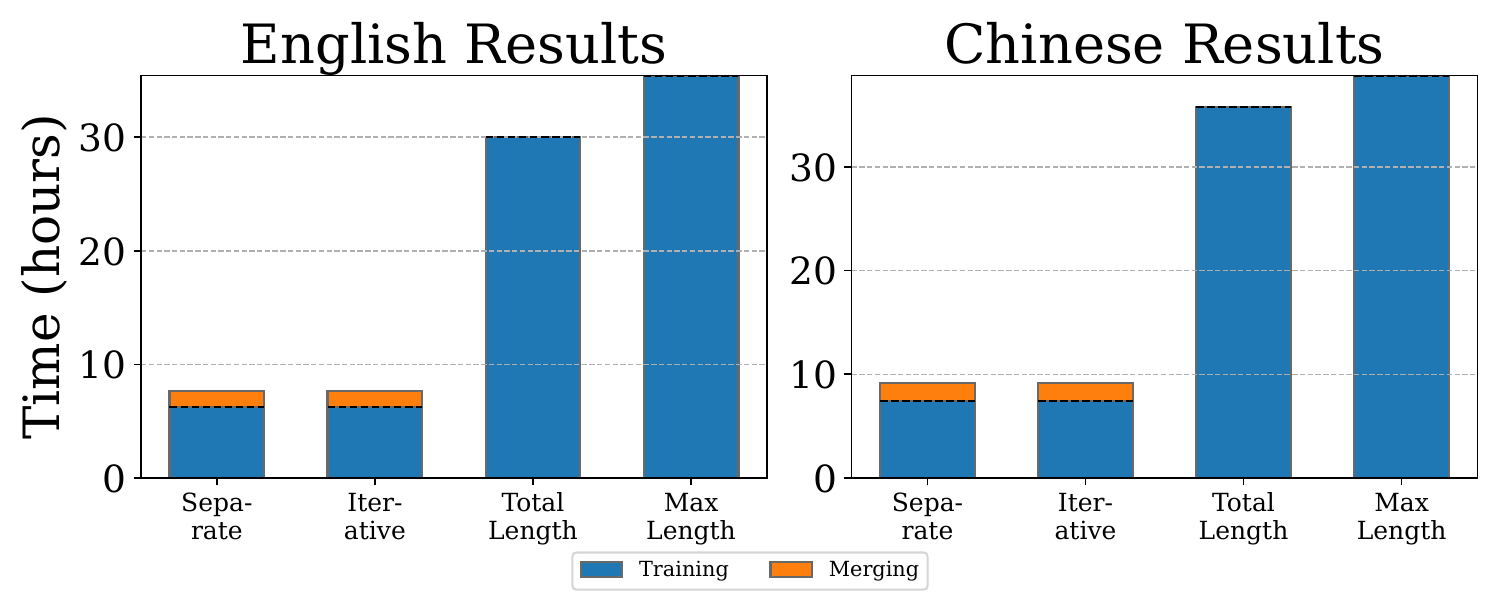}
        \label{fig:tackle_data_imbalance_time_cost}
    }
    \caption{Comparison of Self Positioning (labeled as ``Separate'' and ``Iterative'') and Resampling Methods (``Total Length'' and ``Max Length'') in Tackling Data Imbalance.}
    \label{fig:tackle_data_imbalance}
    \vspace{-2em}
\end{figure}

\subsubsection{Efficiency in Addressing Data Imbalance}
\label{sec:self_positioning_analyses_efficiency}

\begin{table}
  \centering
  \caption{Results on MTEB~\cite{muennighoff_mteb_2023} for Separate Merging(Section ~\ref{sec:applications_separate_merging}) and Iterative Merging(Sectuion ~\ref{sec:applications_iterative_merging}). Retri., Pair., Class., Sum. refer to retrieval, pair classification, classification, and summarization, respectively.  All models utilize the T5-large encoder~\cite{raffel_exploring_2020} as their backbone. Results of GTR and Instructor are sourced from ~\cite{su_one_2023}. 
  For methods requiring sample data (Fisher, RegMean, and our Self Positioning), the first result in each group uses training data, while the second uses alternative data sources. }
  \resizebox{1.0\linewidth}{!}{
    \begin{tabular}{l|cc|cc|cc|cc|cc|cc|cc|cc}
    \toprule
          & \multicolumn{2}{c|}{\textbf{Avg. (56)}} & \multicolumn{2}{c|}{\textbf{Class. (12)}} & \multicolumn{2}{c|}{\textbf{Cluster. (11)}} & \multicolumn{2}{c|}{\textbf{Pair. (3)}} & \multicolumn{2}{c|}{\textbf{ReRank. (4)}} & \multicolumn{2}{c|}{\textbf{Retri. (15)}} & \multicolumn{2}{c|}{\textbf{STS (10)}} & \multicolumn{2}{c}{\textbf{Sum. (1)}} \\
    \midrule
    \textbf{GTR~\cite{ni_large_2022}} & \multicolumn{2}{c|}{58.3} & \multicolumn{2}{c|}{67.1} & \multicolumn{2}{c|}{41.6} & \multicolumn{2}{c|}{85.3} & \multicolumn{2}{c|}{55.4} & \multicolumn{2}{c|}{47.4} & \multicolumn{2}{c|}{78.2} & \multicolumn{2}{c}{29.5} \\
    \textbf{Instructor~\cite{su_one_2023}} & \multicolumn{2}{c|}{61.6} & \multicolumn{2}{c|}{73.9} & \multicolumn{2}{c|}{45.3} & \multicolumn{2}{c|}{85.9} & \multicolumn{2}{c|}{57.5} & \multicolumn{2}{c|}{47.6} & \multicolumn{2}{c|}{83.2} & \multicolumn{2}{c}{31.8} \\
        \midrule
    \multicolumn{17}{c}{\textit{Separate Merging}} \\
    \midrule
    \textbf{TIES} & \multicolumn{2}{c|}{62.2 } & \multicolumn{2}{c|}{72.5 } & \multicolumn{2}{c|}{45.4 } & \multicolumn{2}{c|}{86.3 } & \multicolumn{2}{c|}{57.7 } & \multicolumn{2}{c|}{51.1 } & \multicolumn{2}{c|}{82.8 } & \multicolumn{2}{c}{31.2 } \\
    \textbf{Fisher} & 62.1  & 61.7  & 72.3  & 72.2  & 45.2  & 44.6  & 86.4  & 86.7  & 57.5  & 56.9  & 51.0  & 50.2  & 82.7  & 82.4  & 30.9  & 31.6  \\
    \textbf{RegMean} & 61.5  & 61.4  & 71.9  & 72.0  & 44.7  & 44.5  & 86.6  & 86.7  & 56.9  & 56.8  & 49.7  & 49.6  & 82.4  & 82.4  & 31.6  & 31.5  \\
    \textbf{Self Positioning} & 62.3  & 62.2  & 72.6  & 72.6  & 45.6  & 45.4  & 86.0  & 86.3  & 58.0  & 57.7  & 51.0  & 51.1  & 82.9  & 82.7  & 31.4  & 31.3  \\
    \midrule
    \multicolumn{17}{c}{\textit{Iterative Merging (with extra classification training data) }} \\
    \midrule
    \textbf{TIES} & \multicolumn{2}{c|}{62.8 } & \multicolumn{2}{c|}{78.3 } & \multicolumn{2}{c|}{44.8 } & \multicolumn{2}{c|}{85.1 } & \multicolumn{2}{c|}{57.7 } & \multicolumn{2}{c|}{49.6 } & \multicolumn{2}{c|}{82.2 } & \multicolumn{2}{c}{31.0 } \\
    \textbf{Fisher} & \multicolumn{2}{c|}{62.5 } & \multicolumn{2}{c|}{79.2 } & \multicolumn{2}{c|}{44.5 } & \multicolumn{2}{c|}{84.5 } & \multicolumn{2}{c|}{57.6 } & \multicolumn{2}{c|}{48.7 } & \multicolumn{2}{c|}{81.7 } & \multicolumn{2}{c}{29.9 } \\
    \textbf{RegMean} & \multicolumn{2}{c|}{62.7 } & \multicolumn{2}{c|}{78.8 } & \multicolumn{2}{c|}{44.8 } & \multicolumn{2}{c|}{84.7 } & \multicolumn{2}{c|}{57.4 } & \multicolumn{2}{c|}{49.2 } & \multicolumn{2}{c|}{82.1 } & \multicolumn{2}{c}{30.4 } \\
    \textbf{Self Positioning} & \multicolumn{2}{c|}{63.0 } & \multicolumn{2}{c|}{79.1 } & \multicolumn{2}{c|}{44.6 } & \multicolumn{2}{c|}{85.3 } & \multicolumn{2}{c|}{57.6 } & \multicolumn{2}{c|}{49.9 } & \multicolumn{2}{c|}{82.2 } & \multicolumn{2}{c}{30.5 } \\
    \bottomrule
    \end{tabular}}
  \label{tab:results_mteb}%
\end{table}

In constructing the target dataset $D_\mathrm{t}$, we aim to address the problem of data imbalance by sampling proportionally from both the STS and Retrieval training data. In this section, we compare the approaches of Self Positioning and resampling to tackle data imbalance, focusing on two main dimensions: model performance and time cost. Specifically, we experiment with the ratios of STS training data amount $N_\mathrm{s}$ to Retrieval training data amount $N_\mathrm{r}$, where $\frac{N_\mathrm{s}}{N_\mathrm{r}} \in \{\frac{3}{1}, \frac{3}{2}, \frac{3}{3}, \frac{2}{3}, \frac{1}{3}\}$. 
For Self Positioning, we ensure the total data amount remains constant (at 32,000), and adjust the ratios when constructing the target dataset $D_\mathrm{t}$, while keeping other settings consistent with Section~\ref{sec:self_positioning_experiments}. We apply Self Positioning in both two merging settings (denoted as `Separate' and `Iterative' respectively).
For the resampling method, we consider two scenarios: 1) ensuring the total data amount remains constant (English at 1,040,694, Chinese at 1,185,677), denoted as `Total Length'; 2) keeping the larger dataset's amount constant (English at 769,082, Chinese at 802,291), denoted as `Max Length'. Other settings are maintained as in Section~\ref{sec:problems_setup}. 
The results are illustrated in Figure~\ref{fig:tackle_data_imbalance}. The results show that the Self Positioning method significantly improves both performance and time efficiency.

\section{Applications on Instructor}
\label{sec:applications}


Instructor~\cite{su_one_2023} is an instruction-following embedding model that has been jointly trained on 330 tasks. Our objective is to further validate the proposed merging method by integrating more models. In this section, we apply Self Positioning and the two merging pipelines introduced in Section~\ref{sec:solutions_mergin_pipelines} to Instructor. 

\subsection{Separate Merging}
\label{sec:applications_separate_merging}

In this section, we employ the Separate Merging pipeline to enhance its embedding performance. As there are not all tasks present conflicts, we adopt a clustering approach similar to that in \cite{gou_mixture_2023} to divide the tasks into $N$ clusters. 
For each task cluster, we fine-tune a \verb|gtr-t5-large|~\cite{ni_large_2022} model using the same settings as in ~\cite{su_one_2023}. Subsequently, we apply Self Positioning to explore the merging results. During the search process, we set the batch size to 16 and $\mu \in \{0.00, 0.05\}$, while keeping other settings consistent with those described in Section~\ref{sec:self_positioning_experiments}. To assess the effectiveness of our method beyond the training data, we consider two target dataset settings: 1) constructed by sampling from the Instructor's training data; 2) using the target dataset composed of STS and Retrieval data mentioned in Section~\ref{sec:self_positioning_experiments}. The merged models are evaluated with 56 evaluation tasks in MTEB~\cite{muennighoff_mteb_2023} and the results of 3-cluster are presented in Table~\ref{tab:results_mteb}. Additional results for different cluster numbers and detailed implementation specifics are available in Appendix~\ref{sec:implementation_clustering}.
By comparing the average performance in the table, we can draw the following conclusions: 1) Separate Merging consistently yields significant improvements over the original Instructor model across various settings. This highlights the superiority of Separate Merging compared to joint training. 
2) Analysis of different data sources reveals that Fisher and RegMean rely solely on training data and exhibit substantial performance drops when applied to data from other sources. Conversely, Self Positioning maintains strong performance across diverse data sources, demonstrating its adaptability and flexibility.

\subsection{Iterative Merging}
\label{sec:applications_iterative_merging}

This section attempts to use the Iterative Merging to specifically enhance the performance of the optimal model $M^\mathrm{o}$ in Section~\ref{sec:applications_separate_merging} on particular tasks. 
Specifically, we collect a training dataset for Classification tasks consisting of 209,375 instances (detailed in Appendix~\ref{sec:extra_classification_training_data}) to fine-tune $M^\mathrm{o}$ under the same training setting as in Section~\ref{sec:applications_separate_merging}, with the merged model noted as $M^\mathrm{c}$. Then, we use the search settings from Section~\ref{sec:applications_separate_merging} to search for merging results of $M^\mathrm{o}$ and $M^\mathrm{c}$. Notably, the target dataset $D_\mathrm{t}$ is obtained by sampling from Instructor's training dataset and the training dataset used in this section at a 2:1 ratio. 
The results of our proposed method are compared with other model merging methods in Table~\ref{tab:results_mteb}. 
By comparing the average and classification performance in the table, we can draw the following conclusions: 1) The use of additional classification data combined with Iterative Merging leads to significant improvements in both average and classification performance, demonstrating that Iterative Merging can enhance model performance on certain tasks while maintaining average performance. 2) Self Positioning achieves superior average performance compared to other model merging methods, reflecting the advantages of the Self Positioning method in enhancing performance.

\section{Conclusion}
\label{sec:conclusion}
In this work, we identify two challenges in jointly training general text embeddings that hinder their performance: \textbf{Task Conflict} and \textbf{Data Imbalance}. To address these issues, we propose a model merging-based solution and validate its effectiveness through experiments. Additionally, by analyzing five existing model merging methods used in our experiments, we discover that high-performing merging results can be achieved solely within the interpolation space of task vectors. To mitigate the high time complexity associated with searching for merging results in this space, we introduce an automatic search method called \textbf{Self Positioning}. Through comprehensive experiments and analysis, we demonstrate that Self Positioning effectively searches for optimal merging results and efficiently resolves Data Imbalance. Finally, by applying our proposed method to the Instructor model, we significantly enhance its performance, illustrating the applicability of our approach to mainstream methods.

\bibliography{ref}
\bibliographystyle{unsrt}

\clearpage
\appendix

\section{Experimental Setup}

\subsection{Experimental Environment}

Our experimental environment consists of: \verb|python| 3.10.14, \verb|pytorch| 2.3.0~\cite{paszke_pytorch_2019}, \verb|transformers|~\cite{wolf_transformers_2020} 4.37.2, \verb|mteb|~\cite{muennighoff_mteb_2023} 1.2.0. Experiments with BERT$_\mathrm{base}$~\cite{devlin_bert_2019} are conducted on a single A100 GPU, while experiments on Instructor use two A100 GPUs.

\subsection{Details of Evaluation}
\label{sec:details_of_evaliuation}

Our evaluations utilize MTEB benchmark~\cite{muennighoff_mteb_2023}, which consists of 7 tasks and a total of 56 datasets. The description of each task and its main evaluation matric are as follows:

\begin{itemize}
    \item \textbf{Classification (12 datasets):} Embedding sentences for tasks like sentiment analysis, intent classification. \emph{Main metric: Accuracy}.
    \item \textbf{Clustering (11 datasets):} Clustering sentences or paragraphs using embedded representations. \emph{Main metric: V-measure}.
    \item \textbf{Pair Classification (3 datasets):} Classifying relationships between pairs of texts, such as determining if they are duplicates or paraphrases. \emph{Main metric: Average Precision}.
    \item \textbf{Reranking (4 datasets):} Reranking lists of relevant and irrelevant texts in relation to a query based on embedded similarities. \emph{Main metric: MAP (Mean Average Precision)}.
    \item \textbf{Retrieval (15 datasets):} Retrieving relevant documents for queries from a corpus using embedded representations. \emph{Main metric: nDCG@10 (Normalized Discounted Cumulative Gain)}.
    \item \textbf{Semantic Textual Similarity (STS) (10 datasets):} Determining the similarity between pairs of sentences. \emph{Main metric: Spearman Correlation}.
    \item \textbf{Summarization (1 dataset:} Evaluating the quality of machine-generated summaries compared to human-generated references. \emph{Main metric: Spearman Correlation}.
\end{itemize}

\subsection{Hyperparameter Tuning for Model Merging Methods}

In Section~\ref{sec:solutions_effectiveness}, we conducted a hyperparameter search for each model merging method. Here we present the range of hyperparameter searches for each method:
\begin{itemize}
    \item Average: $(w_\mathrm{s}, w_\mathrm{r}) \in \{(1, 5), (2, 4), (3, 3), (4, 2), (5, 1)\}$;
    \item SLERP: $(a_\mathrm{s}, a_\mathrm{r}) \in \{(1, 5), (2, 4), (3, 3), (4, 2), (5, 1)\}$;
    \item TIES: $\lambda \in \{0.8, 1.0, 1.2, 1.4, 1.6\}$;
    \item Fisher: $(\lambda_\mathrm{s}, \lambda_\mathrm{r}) \in \{(1, 5), (2, 4), (3, 3), (4, 2), (5, 1)\}$;
    \item RegMean: The ratio for reducing non-diagonal elements $\sigma \in \{0.0, 0.1, 0.2, 0.3, 0.4\}$.
\end{itemize}

\section{Task Conflict}

\begin{figure}
    \centering
    \includegraphics[width=.8\columnwidth]{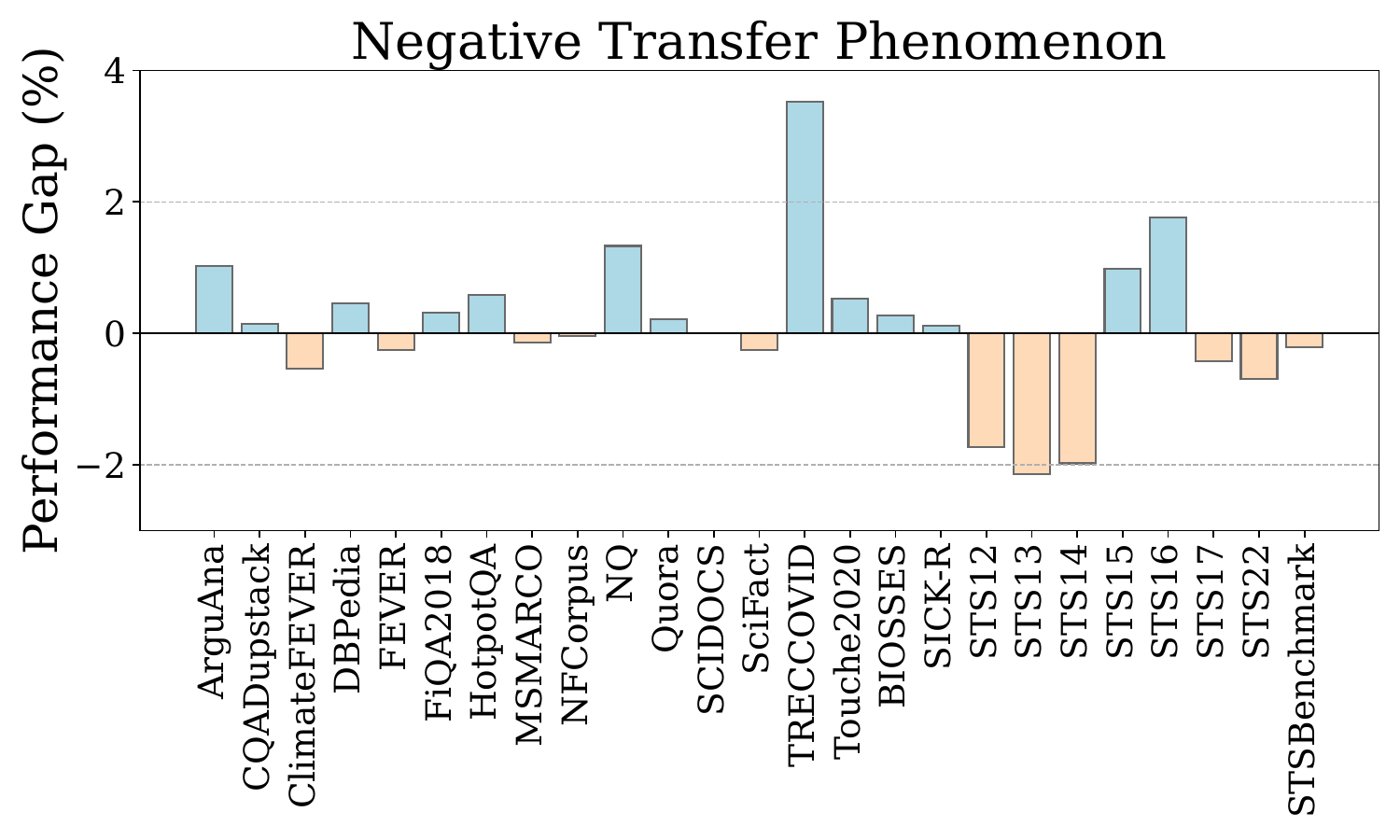}
    \caption{The difference in model performance obtained from joint training compared to the best performance achieved when trained separately on STS and Retrieval tasks in English.}
    \label{fig:problems_task_conflicts_en}
\end{figure}

In the main text, we present the results of task conflict in Chinese; here, we provide the results of task conflict in English in Figure~\ref{fig:problems_task_conflicts_en}. From these results, it is evident that although the Negative Transfer Phenomenon is less pronounced in English than in Chinese, there is still a decline in performance in 12 out of the 25 tasks (nearly half).

\section{Loss Landscape}
\label{sec:loss_landscape}

When computing the loss landscape, we first measure the angle $\alpha$ between the task vectors $V_\mathrm{s}$ and $V_\mathrm{r}$. We then represent the backbone model parameters in the bottom left corner of the image and symmetrically distribute $V_\mathrm{s}$ and $V_\mathrm{s}$ on either side of the image diagonal based on the magnitude of $\alpha$. Next, we measure the magnitudes of the two task vectors, $|V_\mathrm{s}|$ and $|V_\mathrm{r}|$, and set the longer task vector's value on the x-axis (or y-axis) of the image to 0.7, scaling the shorter vector proportionally, to obtain the positions $p_\mathrm{s}$ and $p_\mathrm{r}$ of the two task vectors in the image. We then divide the image into a 20x20 grid, and each grid cell calculates its interpolation weight from the $p_\mathrm{s}$ and $p_\mathrm{r}$, which is used as the model merging weight. We then measure the merged model's loss on a multi-task dataset, which represents the loss of one point on the loss landscape. For computing the loss, we sample data proportionally from the STS and Retrieval training data, calculate the loss with a batch size of 32, and ultimately compile the average result over 200 steps.

\section{Implementation of Clustering}
\label{sec:implementation_clustering}

In Section~\ref{sec:applications_separate_merging}, we divide 330 tasks using a clustering algorithm. Specifically, we sample 100 instances each of $q$ and $p$ from every dataset, and then we obtain their embeddings using the backbone (i.e., \verb|gtr-t5-large|). Next, we perform average pooling on the embeddings of $q$ and $p$ separately and concatenate them to form the dataset's corresponding embedding. We then use sklearn's~\cite{pedregosa_scikit-learn_2011} KMeans algorithm with the following parameters to perform clustering: \verb|'init'='k-means++'|, \verb|'n_init'=10|, \verb|'max_iter'=300|. The final clustering statistics are shown in Table~\ref{tab:statistics_clustering}. The results of varying numbers of clusters are shown in Table~\ref{tab:appendix_results_mteb}. 

\begin{table}
  \centering
  \caption{Statistics for the clustering results of 330 tasks.}
    \begin{tabular}{c|ll}
    \toprule
          & \multicolumn{1}{c}{\textbf{Task Count}} & \multicolumn{1}{c}{\textbf{Instance Count}} \\
    \midrule
    \textbf{2 clusters} & 30/300 & 1,255,000/18,000 \\
    \textbf{3 clusters} & 30/121/179 & 1,255,000/72,600/107,400 \\
    \textbf{4 clusters} & 30/92/97/111 & 1,255,000/55,200/58,200/66,600 \\
    \bottomrule
    \end{tabular}%
  \label{tab:statistics_clustering}%
\end{table}%

\begin{table*}
  \centering
  \caption{Results on MTEB~\cite{muennighoff_mteb_2023}. Retri., Pair., Class., Sum. refer to retrieval, pair classification, classification, and summarization, respectively.  All models utilize the T5-large encoder~\cite{raffel_exploring_2020} as their backbone. Results of GTR and Instructor are sourced from ~\cite{su_one_2023}. Results from Separate Merging (Section~\ref{sec:applications_separate_merging}) are categorized under ``2 clusters'', ``3 clusters'', and ``4 clusters''. 
  For methods requiring sample data (Fisher, RegMean, and our Self Positioning), the first result in each group uses training data, while the second uses alternative data sources. }
  \resizebox{1.0\linewidth}{!}{
    \begin{tabular}{l|cc|cc|cc|cc|cc|cc|cc|cc}
    \toprule
          & \multicolumn{2}{c|}{\textbf{Avg. (56)}} & \multicolumn{2}{c|}{\textbf{Class. (12)}} & \multicolumn{2}{c|}{\textbf{Cluster. (11)}} & \multicolumn{2}{c|}{\textbf{Pair. (3)}} & \multicolumn{2}{c|}{\textbf{ReRank. (4)}} & \multicolumn{2}{c|}{\textbf{Retri. (15)}} & \multicolumn{2}{c|}{\textbf{STS (10)}} & \multicolumn{2}{c}{\textbf{Sum. (1)}} \\
    \midrule
    \textbf{GTR~\cite{ni_large_2022}} & \multicolumn{2}{c|}{58.3} & \multicolumn{2}{c|}{67.1} & \multicolumn{2}{c|}{41.6} & \multicolumn{2}{c|}{85.3} & \multicolumn{2}{c|}{55.4} & \multicolumn{2}{c|}{47.4} & \multicolumn{2}{c|}{78.2} & \multicolumn{2}{c}{29.5} \\
    \textbf{Instructor~\cite{su_one_2023}} & \multicolumn{2}{c|}{61.6} & \multicolumn{2}{c|}{73.9} & \multicolumn{2}{c|}{45.3} & \multicolumn{2}{c|}{85.9} & \multicolumn{2}{c|}{57.5} & \multicolumn{2}{c|}{47.6} & \multicolumn{2}{c|}{83.2} & \multicolumn{2}{c}{31.8} \\
    \midrule
    \multicolumn{17}{c}{\textit{2 clusters}} \\
    \midrule
    \textbf{TIES} & \multicolumn{2}{c|}{62.3 } & \multicolumn{2}{c|}{72.8 } & \multicolumn{2}{c|}{45.7 } & \multicolumn{2}{c|}{86.2 } & \multicolumn{2}{c|}{57.8 } & \multicolumn{2}{c|}{50.7 } & \multicolumn{2}{c|}{83.1 } & \multicolumn{2}{c}{31.3 } \\
    \textbf{Fisher} & 62.3  & 62.1  & 72.8  & 72.7  & 45.4  & 45.1  & 86.1  & 86.4  & 57.8  & 57.6  & 51.2  & 50.8  & 82.9  & 82.9  & 30.8  & 31.0  \\
    \textbf{RegMean} & 61.9  & 61.9  & 72.4  & 72.5  & 45.2  & 45.2  & 86.5  & 86.6  & 57.3  & 57.3  & 50.1  & 50.0  & 83.0  & 83.0  & 31.1  & 31.0  \\
    \textbf{Self Positioning} & 62.2  & 62.3  & 72.8  & 72.8  & 45.7  & 45.5  & 85.9  & 86.3  & 57.8  & 57.7  & 50.3  & 50.8  & 83.1  & 83.0  & 31.3  & 30.9  \\
    \midrule
    \multicolumn{17}{c}{\textit{3 clusters}} \\
    \midrule
    \textbf{TIES} & \multicolumn{2}{c|}{62.2 } & \multicolumn{2}{c|}{72.5 } & \multicolumn{2}{c|}{45.4 } & \multicolumn{2}{c|}{86.3 } & \multicolumn{2}{c|}{57.7 } & \multicolumn{2}{c|}{51.1 } & \multicolumn{2}{c|}{82.8 } & \multicolumn{2}{c}{31.2 } \\
    \textbf{Fisher} & 62.1  & 61.7  & 72.3  & 72.2  & 45.2  & 44.6  & 86.4  & 86.7  & 57.5  & 56.9  & 51.0  & 50.2  & 82.7  & 82.4  & 30.9  & 31.6  \\
    \textbf{RegMean} & 61.5  & 61.4  & 71.9  & 72.0  & 44.7  & 44.5  & 86.6  & 86.7  & 56.9  & 56.8  & 49.7  & 49.6  & 82.4  & 82.4  & 31.6  & 31.5  \\
    \textbf{Self Positioning} & 62.3  & 62.2  & 72.6  & 72.6  & 45.6  & 45.4  & 86.0  & 86.3  & 58.0  & 57.7  & 51.0  & 51.1  & 82.9  & 82.7  & 31.4  & 31.3  \\
    \midrule
    \multicolumn{17}{c}{\textit{4 clusters}} \\
    \midrule
    \textbf{TIES} & \multicolumn{2}{c|}{62.1 } & \multicolumn{2}{c|}{72.3 } & \multicolumn{2}{c|}{45.2 } & \multicolumn{2}{c|}{86.4 } & \multicolumn{2}{c|}{57.6 } & \multicolumn{2}{c|}{51.2 } & \multicolumn{2}{c|}{82.6 } & \multicolumn{2}{c}{31.3 } \\
    \textbf{Fisher} & 62.1  & 61.7  & 72.3  & 72.2  & 45.2  & 44.6  & 86.4  & 86.7  & 57.5  & 56.9  & 51.0  & 50.2  & 82.7  & 82.4  & 30.9  & 31.6  \\
    \textbf{RegMean} & 61.5  & 61.4  & 71.9  & 72.0  & 44.7  & 44.5  & 86.6  & 86.7  & 56.9  & 56.8  & 49.7  & 49.6  & 82.4  & 82.4  & 31.6  & 31.5  \\
    \textbf{Self Positioning} & 62.3  & 62.2  & 72.6  & 72.6  & 45.6  & 45.4  & 86.0  & 86.3  & 58.0  & 57.7  & 51.0  & 51.1  & 82.9  & 82.7  & 31.4  & 31.3  \\
    \midrule
    \multicolumn{17}{c}{\textit{Iterative Merging (with extra classification training data) }} \\
    \midrule
    \textbf{TIES} & \multicolumn{2}{c|}{62.8 } & \multicolumn{2}{c|}{78.3 } & \multicolumn{2}{c|}{44.8 } & \multicolumn{2}{c|}{85.1 } & \multicolumn{2}{c|}{57.7 } & \multicolumn{2}{c|}{49.6 } & \multicolumn{2}{c|}{82.2 } & \multicolumn{2}{c}{31.0 } \\
    \textbf{Fisher} & \multicolumn{2}{c|}{62.5 } & \multicolumn{2}{c|}{79.2 } & \multicolumn{2}{c|}{44.5 } & \multicolumn{2}{c|}{84.5 } & \multicolumn{2}{c|}{57.6 } & \multicolumn{2}{c|}{48.7 } & \multicolumn{2}{c|}{81.7 } & \multicolumn{2}{c}{29.9 } \\
    \textbf{RegMean} & \multicolumn{2}{c|}{62.7 } & \multicolumn{2}{c|}{78.8 } & \multicolumn{2}{c|}{44.8 } & \multicolumn{2}{c|}{84.7 } & \multicolumn{2}{c|}{57.4 } & \multicolumn{2}{c|}{49.2 } & \multicolumn{2}{c|}{82.1 } & \multicolumn{2}{c}{30.4 } \\
    \textbf{Self Positioning} & \multicolumn{2}{c|}{63.0 } & \multicolumn{2}{c|}{79.1 } & \multicolumn{2}{c|}{44.6 } & \multicolumn{2}{c|}{85.3 } & \multicolumn{2}{c|}{57.6 } & \multicolumn{2}{c|}{49.9 } & \multicolumn{2}{c|}{82.2 } & \multicolumn{2}{c}{30.5 } \\
    \bottomrule
    \end{tabular}}
  \label{tab:appendix_results_mteb}%
\end{table*}%

\section{Extra Classification Training Data}
\label{sec:extra_classification_training_data}

We use the English training portions of various classification datasets from the MTEB~\cite{muennighoff_mteb_2023}. The specific datasets used include: AmazonReviews Classification~\cite{DBLP:conf/recsys/McAuleyL13}, AmazonCounterfactual Classification~\cite{DBLP:conf/emnlp/ONeillRKKB21}, Banking77-Classification~\cite{DBLP:journals/corr/abs-2003-04807}, Emotion-Classification~\cite{DBLP:conf/emnlp/SaraviaLHWC18}, IMDB- Classification~\cite{DBLP:conf/acl/MaasDPHNP11}, MTOPIntent-Classification~\cite{DBLP:conf/eacl/LiACGGM21}, ToxicConversations Classification, and TweetSentimentExtraction Classification. We limit the maximum instance number of all datasets to 25,000.

\end{document}